\documentclass[11pt]{article}
\usepackage{paclic35}
\usepackage{times}
\usepackage{latexsym}
\usepackage{amsmath}
\usepackage{multirow}
\usepackage{url}

\usepackage[utf8]{inputenc}

\usepackage{danudefs}
\usepackage{algorithmic}
\usepackage{algorithm}
\usepackage{color}
\usepackage{booktabs,arydshln}
\usepackage{xspace}
\usepackage{url}
\usepackage{graphicx}
\usepackage{subfigure} 
\usepackage[section]{placeins}
\usepackage{xcolor}
\usepackage{microtype}

\usepackage[english]{babel}
\addto\captionsenglish{%
}
\addto\extrasenglish{%
}

\makeatletter
\def\adl@drawiv#1#2#3{%
        \hskip.5\tabcolsep
        \xleaders#3{#2.5\@tempdimb #1{1}#2.5\@tempdimb}%
                #2\z@ plus1fil minus1fil\relax
        \hskip.5\tabcolsep}
\newcommand{\cdashlinelr}[1]{%
  \noalign{\vskip\aboverulesep
           \global\let\@dashdrawstore\adl@draw
           \global\let\adl@draw\adl@drawiv}
  \cdashline{#1}
  \noalign{\global\let\adl@draw\@dashdrawstore
           \vskip\belowrulesep}}
\makeatother

\DeclareMathOperator*{\avg}{avg}
\usepackage{array}
\newcolumntype{L}[1]{>{\raggedright\let\newline\\\arraybackslash\hspace{0pt}}m{#1}}

\title{Learning Sense-Specific Static Embeddings using \\Contextualised Word Embeddings as a Proxy}
  
\author{Yi Zhou \\
  Department of Computer Science \\ University of Liverpool, UK \\
  {\tt Y.Zhou71@liverpool.ac.uk} \\\And
  Danushka Bollegala \\
  Department of Computer Science \\ University of Liverpool, UK \\
  {\tt danushka@liverpool.ac.uk} \\}

\date{}

\begin{document}
\maketitle

\begin{abstract}
Contextualised word embeddings generated from Neural Language Models (NLMs), such as BERT, represent a word with a vector that considers the semantics of the target word as well its context.
On the other hand, static word embeddings such as GloVe represent words by relatively low-dimensional, memory- and compute-efficient vectors but are not sensitive to the different senses of the word.
We propose Context Derived Embeddings of Senses (CDES), a method that extracts sense related information from contextualised embeddings and injects it into static embeddings to create sense-specific static embeddings.
Experimental results on multiple benchmarks for word sense disambiguation and sense discrimination tasks show that CDES can accurately learn sense-specific static embeddings reporting comparable performance to the current state-of-the-art sense embeddings.
\end{abstract}

\section{Introduction}

Representing the meanings of words using low-dimensional vector embeddings has become a standard technique in NLP. 
\emph{Static word embeddings}~\cite{mikolov2013distributed,pennington2014glove} represent words at the \emph{form} level by assigning a single vector for all occurrences of a word irrespective of its \emph{senses}. 
However, representing ambiguous words such as \emph{bass}, which could mean either a \emph{musical instrument} or a type of \emph{fish}, using a single embedding is problematic.

To address this problem, \emph{sense-specific static word embedding} methods~\cite{reisinger2010multi,neelakantan2014efficient,huang2012improving} assign multiple embeddings to a single polysemous word corresponding to its multiple senses.
However, these embeddings are context-insensitive and we must resort to different heuristics such as selecting the sense embedding of the ambiguous word that is most similar to the context, to determine which embedding should be selected to represent the word.

\begin{figure}[t]
\centering
\includegraphics[width=0.5\textwidth]{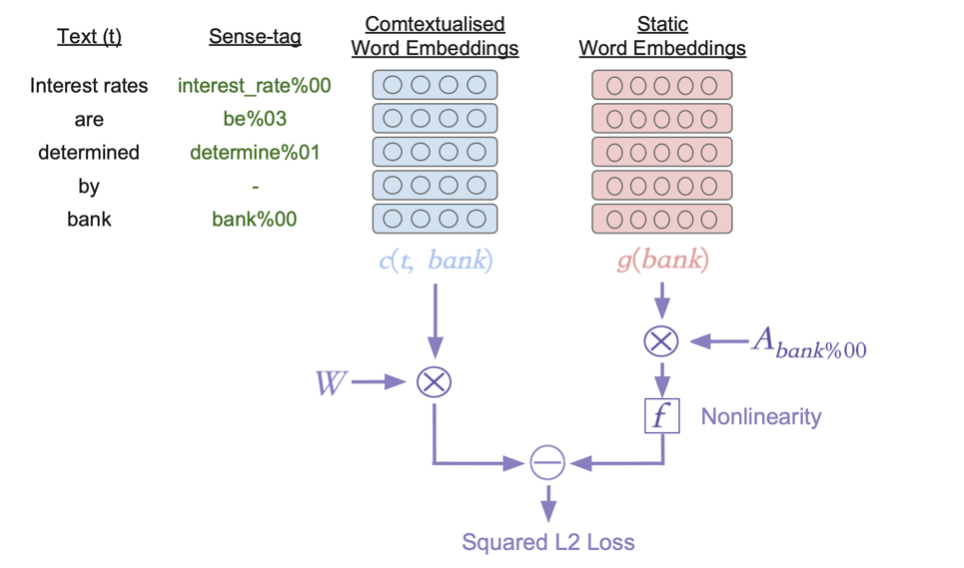}
\caption{Outline of CDES. Given a sense-tagged sentence $t$, we compute a sense embedding for the ambiguous word \emph{bank} by multiplying its static word embedding, $\vec{g}(\textit{bank})$, by a sense-specific projection matrix, $\mat{A}_{\textit{bank\%00}}$, corresponding to the correct sense of the word. Projection matrices are learnt by minimising the squared $\ell_2$ loss between the linearly transformed (via a matrix $\mat{W}$) contextualised embedding, $\vec{c}(t, \textit{bank})$, and of the (nonlinearly transformed via function $f$) sense embedding of \emph{bank}.}
\label{fig:outline}
\end{figure}

On the other hand, \emph{contextualised word embeddings} generated from NLMs~\cite{Elmo,devlin2019bert,liu2019roberta} represent a word in a given context by an embedding that considers both the meaning of the word itself as well as its context.
Different types of information such as word sense, dependency, and numeracy have shown to be encoded in contextualised word embeddings, providing rich, context-sensitive input representations for numerous downstream NLP applications.
More recently, \newcite{loureiro2019language} and \newcite{scarlini2020sensembert} showed that contextualised embeddings such as BERT~\cite{devlin2019bert} and ELMo~\cite{Elmo} can be used to create sense embeddings by means of external semantic networks, such as WordNet~\cite{miller1998wordnet} and BabelNet~\cite{navigli2010babelnet}. 
Moreover, \newcite{levine-etal-2020-sensebert} showed that BERT can be fine-tuned using WordNet's supersenses to learn contextualised sense embeddings.

Inspired by these prior successes, we ask and affirmatively answer the question -- \emph{can we extract sense-related information from contextualised word embeddings to create sense-specific versions of (pretrained) sense-agnostic static embeddings?}
To this end, we propose, Context Derived Embeddings of Senses (\textbf{CDES})
a method to extract sense-related information encoded in contextualised word embeddings and inject it into pretrained sense-agnostic static word embeddings to create sense-specific static embeddings.
Given a contextualised embedding, a static word embedding and a sense-annotated corpus, CDES learns sense-specific projection matrices that can be used to predict the sense embeddings of words from their word embeddings.
Following the distributional hypothesis~\cite{DS}, we require that the predicted sense embedding of a word must align (possibly nonlinearly) with the meaning of the context of the word, represented using a contextualised embedding as outlined in Figure~\ref{fig:outline}.

At a more conceptual level, CDES can be seen as using contextualised language models as a proxy for extracting information relevant to a particular task, without learning it directly from text corpora. 
In particular, prior work probing language models has shown that rich information about languages is compactly and accurately encoded within contextualised representations produced by NLMs~\cite{klafka-ettinger-2020-spying}.
Moreover, CDES can also be seen as an instance of \emph{model distillation}~\cite{Furlanello:2018aa}, where a complex teacher model (i.e. a contextualsied word embedding) is used to train a simpler student model (i.e. a sense-sensitive static embedding).

There are several advantages in CDES for learning sense-specific static embeddings. 
CDES is computationally relatively lightweight because it uses \emph{pretrained} static embeddings as well as contextualised embeddings from a \emph{pretrained} NLM and does not require training these resources from scratch. 
CDES static sense embeddings can be precomputed because of their independence on the context.
Therefore, CDES embeddings are attractive to NLP applications that must run on limited hardware resources. 
Because subtokenisation methods, such as Byte Pair Encoding (BPE), must be used to limit the vocabulary sizes, one must post-process subtoken embeddings (e.g. by mean pooling) to create word embeddings with contextualised embeddings, whereas static embeddings can directly learn word embeddings. 
To increase the coverage of sense embeddings, in addition to the sense related information extracted from contextualised embeddings, CDES incorporates contextual information from external corpora and knowledge bases.

We evaluate CDES on Word Sense Disambiguation~\cite[WSD]{navigli2009word} (Section~\ref{sec:WSD}) and Words in Context~\cite[WiC]{Pilehvar:2019} (Section~\ref{sec:wic}) tasks.
In both tasks, CDES learns accurate sense embeddings and outperforms many existing static sense embeddings.
In particular, on the WSD framework~\cite{raganato2017word}, CDES reports the best performance in 4 out of 6 benchmarks,
and on WiC reports competitive results to the current state-of-the-art without any fine-tuning of on task data.

\section{Context-Derived Embedding of Senses}
\label{sec:method}

Given (a) pretrained static word embeddings, (b) contextualised word embeddings from a pretrained NLM, and (c) a sense-annotated corpus, CDES  learns a sense-specific version of (a), representing each sense of a word by a different vector.
To describe CDES in detail, let us denote the sense-agnostic static embedding of a word  $u \in \cV$ in a vocabulary $\cV$, by $\vec{g}(u) \in \R^{p}$.
Moreover, let us denote the contextualised embedding model $c$, from which we can obtain a context-sensitive representation $\vec{c}(u, t) \in \R^{q}$ corresponding to $u$ in some context $t \in \cC(u)$.
Here, $\cC(u)$ is the set of contexts in which $u$ occurs.
An ambiguous word $u$ is likely to take different senses in different contexts $t$, and our goal is to learn a sense-specific embedding of $u$ that captures the different senses of $u$.

Let us denote by $\cS$ the set of word senses taken by all words in $\cV$.
An ambiguous word $u$ will belong to a subset $\cS(u)$ of senses in $\cS$.
Let us denote the sense-specific embedding of $u$ corresponding to the $i$-th sense $s_{i} \in \cS(u)$ by $\vec{s}_{i}(u) \in \R^{p}$.
We model the process of creating sense-specific embeddings from static embeddings as a projection learning task, where we multiply the static embedding, $\vec{g}(u)$, by a sense-specific projection matrix, $\mat{A}_{i}$, to produce $\vec{s}_{i}(u)$ as in \eqref{eq:proj}.
\begin{align}
 \label{eq:proj}
 \vec{s}_{i}(u) = \mat{A}_{i} \vec{g}(u)
\end{align}
Here, \eqref{eq:proj} decouples a sense embedding into a sense-agnostic static lexical semantic component given by $\vec{g}(u)$ and 
a word-independent sense-specific component $\mat{A}_{i}$, enabling efficient sense-specific embedding learning using pretrained embeddings.
The projection matrices can be seen as linear operators that produce different sense-specific embeddings from the same static word (lemma) embedding, corresponding to the different senses of the lemma.

On the other hand $\vec{c}(u,t)$ encodes both sense related information for $u$ as well as information not related to $u$ such as the grammatical gender or number in the context $t$.
Therefore, we apply a linear filter parameterised by a matrix $\mat{W} \in \R^{q \times p}$, to extract sense related information from $\vec{c}(u,t)$. 

Given a sense tagged corpus, we jointly learn $\mat{W}$ and $\mat{A}_{i}$s by minimising the objective given by \eqref{eq:loss}.
\par\nobreak
{\small
\begin{align}
\label{eq:loss}
L(\mat{W}, \{\mat{A}_i\}_{i=1}^{|\cS|}) =  \sum_{\substack{u \in \cV \\  t \in \cC(u) \\ s_i \in \cS(u)}} \norm{\mat{W} \vec{c}(u,t) - f(\mat{A}_i\vec{g}(u))}_2^2
\end{align}
}%




Here, $f$ is an elementwise nonlinear function that enables us to consider nonlinear associations between contextualised and static word embeddings.
In our experiments, we consider linear, ReLU and GELU activations as $f$.
After training, we can compute the sense embeddings $\vec{s}_{i}(u)$ using \eqref{eq:proj} with the pretrained static word embeddings $\vec{g}(u)$.

Eq.~\eqref{eq:loss} can be seen as aligning the contextualised and static word embeddings under a nonlinear transformation. 
The only learnable parameters in our proposed method are $\mat{W}$ and sense-specific projections $\mat{A}_{1}, \ldots, \mat{A}_{|\cS|}$.
In particular, we \emph{do not} require re-training or fine-tuning static or contextualised embeddings
and can be seen as a post-processing method applied to pretrained embeddings, similar to retrofitting~\cite{shi-etal-2019-retrofitting}.
We limit the sense-specific projection matrices to diagonal matrices in our experiments because in our preliminary investigations we did not find any significant advantage in using full matrices compared to the extra storage.
Moreover, a diagonal matrix can be compactly represented by storing only its diagonal elements as a vector, which reduces the number of parameters to learn (thus less likely to overfit) and speeds up matrix-vector multiplications.

\subsection{Context Aggregation}
\label{sec:context}
An important limitation of the above-mentioned setting is that it requires sense-annotated corpora.
Manually annotating word senses in large text corpora is expensive and time consuming.
Moreover, such resources might not be available for low resource languages.
Even if such sense-annotated corpora are available for a particular language, they might not cover all different senses of all of the words in that language, resulting in an inadequate sense coverage.
For example, SemCor~\cite{SemCor}, one of the largest manually-annotated corpora for English word senses including more than 220K words tagged with 25K distinct WordNet meanings, covers only 15\% of all synsets in the WordNet.
To address this sense-coverage problem, we follow prior proposals~\cite{scarlini2020more} to extract additional contexts for a word from (a) \textbf{the dictionary definitions of synsets}, and (b) \textbf{an external corpus}.

\paragraph{Gloss-based Sense Embeddings:}
To create sense embeddings from dictionary definitions, we use the glosses of synsets in the WordNet.
Given a word $u$, we create a gloss-based sense embedding, $\vec{\psi}(u)_i \in \R^q$, represented by the sentence embedding, $\vec{c}(t_i)$, computed from the gloss $t_i$ corresponding to the synset $s_i$ of $u$.
Here, $\vec{c}(t_i)$ is computed by averaging the contextualised embeddings for the tokens in the gloss $t_i$ as given in \eqref{eq:context}.
\begin{align}
    \label{eq:context}
    \vec{c}(t_i) = \avg_{w \in t_i}\vec{c}(w,t_i)
\end{align}
Here, $\mathrm{avg}$ denotes mean pooling over the tokens $w$ in $t_i$.
Following \newcite{loureiro2019language} and \newcite{scarlini2020more}, in our experiments, we use BERT as the contextualised embedding model and use the sum of the final four layers as token embeddings.

\paragraph{Corpus-based Sense Embeddings:}
To extract contexts from an external corpus for given a word $u$, we retrieve all sentences as contexts $t \in \cC(u)$ from the corpus where $u$ occurs.
We then cluster the extracted sentences (represented by the sentence embeddings computed using \eqref{eq:context}) using the $k$-means algorithm.
We assume each cluster to contain similar sentences and that $u$ will be used in the same sense in all sentences in a cluster.
We use UKB\footnote{\url{http://ixa2.si.ehu.eus/ukb/}}~\cite{agirre-etal-2014-random}, a knowledge-based approach to WSD that uses the Personalised PageRank algorithm~\cite{Haveliwala_2002}, to disambiguate the clusters.

To increase the coverage of senses represented by the clusters, we consider collocations of $u$ available in SyntagNet~\cite{maru-etal-2019-syntagnet}\footnote{\url{http://syntagnet.org/}} following \newcite{scarlini2020more}. 
Specifically, for each word $u$, we find words $v$ that forms a collocation with $u$ in SyntagNet, and extract sentences $t$ that contain both $u$ and $v$ within a co-occurrence window.
The synset id $s_i$ assigned to the $(u,v)$ pair in SyntagNet is used as the sense id for all extracted sentences for $u$.
Finally, we compute a corpus-based sense embedding $\vec{\phi}_i(u) \in \R^q$ as the cluster centroid, where sentence embeddings are computed using \eqref{eq:context}.

\subsection{Sense Embedding and Disambiguation}
\label{sec:sense}

The final CDES static sense embedding, $\mathbf{cdes}_i(u) \in \R^{p+2q}$ of the $i$-th sense of $u$ is computed as the concatenation of
$\vec{s}_i(u)$ (given by \eqref{eq:proj}), gloss-based sense embedding $\vec{\psi}_i(u)$ and corpus-based sense embedding $\vec{\phi}_i(u)$ as given by \eqref{eq:cdes}, where $\oplus$ denotes vector concatenation.
\begin{align}
    \label{eq:cdes}
    \mathbf{cdes}_i(u) = \vec{s}_i(u) \oplus \vec{\psi}_i(u) \oplus \vec{\phi}_i(u)
\end{align}

In order to disambiguate a word $u$ in a given context $t'$, we first compute a contextualised embedding $\vec{\zeta}(u,t') \in \R^{p+2q}$ by concatenating three vectors as give by \eqref{eq:wsd}
\begin{align}
    \label{eq:wsd}
    \vec{\zeta}(u,t') = \vec{g}(u) \oplus \vec{c}(u,t') \oplus \vec{c}(u,t')
\end{align}
We then compute the cosine similarity between $\vec{\zeta}(u,t')$ and $\mathbf{cdes}_i(u)$ for each sense $s_i$ of $u$.
We limit the candidate senses based on the lemma and part-of-speech of $u$ in $t'$, and select the most similar (1-NN) sense of $u$ as its disambiguated sense in context $t'$.

\section{Experiments}

\subsection{Experimental Setup}

In our experiments, we use the pretained GloVe\footnote{\url{nlp.stanford.edu/projects/glove/}} embeddings (Common Crawl with 840B tokens and 2.2M vocabulary) as the static word embeddings $\vec{g}(u)$ with $p = 300$.
We use pretrained BERT (\texttt{large-bert-cased}\footnote{\url{https://bit.ly/33Nsmou}}) as the contextualised embedding model, $\vec{c}(u,t)$ with $q = 1024$.
Following prior work~\cite{luo2018leveraging,luo2018incorporating,loureiro2019language,scarlini2020more}, we use sense annotations from SemCor $3.0$~\cite{SemCor} as the sense-tagged corpus, which is the largest corpus annotated with WordNet sense ids.
As the external corpus for extracting contexts as described in Section~\ref{sec:context}, we use the English Wikpedia.
The number of clusters in $k$-means is set to the number of distinct senses for the lexeme according to the WordNet.
The number of words given to UKB is set to 5 and the number of sentences extracted from Wikipedia per lemma is set to 150 following \newcite{scarlini2020more}.
The co-occurrence window size for considering collocations extracted from SyntagNet is set to 3 according to \newcite{maru-etal-2019-syntagnet}.
We evaluate the learnt sense embeddings in two downstream tasks: WSD (Section~\ref{sec:WSD}) and WiC (Section~\ref{sec:wic}).
The statistics of SemCor, all-words English WSD and WiC datasets are showed in Table~\ref{tbl:statistics}.

\begin{table}[t]
\centering
\resizebox{0.48\textwidth}{!}{
\begin{tabular}{lccccc}
\toprule
Dataset &Total &Nouns &Vebs &Adj &Adv \\
\midrule
SemCor &226,036 &87,002 &88,334 &31,753 &18,947 \\
\midrule
\textbf{WSD}& & & & & \\
SE2 &2,282 &1,066 &517 &445 &254 \\
SE3 &1,850 &900 &588 &350 &12 \\
SE07 &455 &159 &296 &- &- \\
SE13 &1,644 &1,644 &- &- &- \\
SE15 &1,022 &531 &251 &160 &80 \\
ALL &7,253 &4,300 &1,652 &955 &346 \\
\midrule
\midrule
\textbf{WiC} &Instances &Nouns &Vebs &\multicolumn{2}{c}{Unique Words} \\
\midrule
Training &5,428 &2,660 &2,768 &\multicolumn{2}{c}{1,256} \\
Dev &638 &396 &242 &\multicolumn{2}{c}{599} \\
Test &1,400 &826 &574 &\multicolumn{2}{c}{1,184} \\
\bottomrule
\end{tabular}}
    \caption{The statistics of the training and evaluation datasets. SemCor  is used for training. SemEval (SE07, SE13, SE15) and Senseval (SE2, SE3) datasets are used for the WSD task, whereas the WiC dataset is used for sense discrimination task.}
\label{tbl:statistics}
\vspace{-5mm}
\end{table}

To project contextualised and static word embeddings to a common space, we set $\mat{W} \in \R^{300 \times 1024}$.
To reduce the memory footprint, number of trainable parameters and thereby overfitting, we constrain the sense-specific matrices $\mat{A}_i \in \R^{300 \times 300}$ to be diagonal.
We initialise all elements of $\mat{W}$ and $\mat{A}_i$s uniformly at random in $[0,1]$.
We use Adam as the SGD optimiser and set the minibatch size to $64$ with an initial learning rate of 1E-4. 
All hyperparameter values were tuned using a randomly selected subset of training data set aside as a validation dataset.
The t-SNE visualisations in the paper are produced with \texttt{sklearn.manifold.TSNE} using n\_components=2, init=\emph{pca}, perplexity=3, n\_iter=1500 and metric=\emph{cosine}.

All experiments were conducted on a machine with a single Titan V GPU (12 GB RAM), Intel Xeon 2.60 GHz CPU (16 cores) and 64 GB of RAM.
Overall, training time is less than 3 days on this machine,

\subsection{Word Sense Disambiguation (WSD)}
\label{sec:WSD}

%

WSD is a fundamental task in NLP, which aims to identify the exact sense of an ambiguous word in a given context~\cite{navigli2009word}.  
To evaluate the proposed sense embeddings, we conduct a WSD task using the evaluation framework proposed by~\newcite{raganato2017word}, which includes all-words English WSD datasets: Senseval-2 (SE2), Senseval-3 task 1 (SE3), SemEval-07 task 17 (SE07), SemEval-13 task 12 (SE13) and SemEval-15 task 13 (SE15).
We used the framework's official scoring scripts to avoid any discrepancies in the scoring methodology.
As described in Section~\ref{sec:sense}, the sense of a word in a context is predicted by the 1-NN method.

Table~\ref{tbl:wsd} shows the WSD results. 
Most Frequent Sense (MFS) baseline selects the most frequent sense of a word in the training corpus and has proven to be a strong baseline~\cite{McCarthy:2007}.
\newcite{scarlini2020more} use \newcite{Elmo}'s method with BERT on SemCor+OMSTI~\cite{taghipour2015one} to propose SemCor+OMSTI$_{BERT}$ baseline.
ELMo $k$-NN uses ELMo embeddings to predict the sense of a word following the nearest neighbour strategy.
Specifically, they first obtain ELMo embeddings for all words in SemCor sentences, and average the embeddings for each sense.
At test time, they run ELMo on the given test sentence containing the ambiguous word and select the sense with the highest cosine similarity. 
\newcite{loureiro2019language} repeated this method using BERT~\cite{devlin2019bert} embeddings to propose the BERT $k$-NN baseline.
EWISE$_{ConvE}$~\cite{kumar2019zero} learns a sentence encoder for sense definition by using WordNet relations as well as ConvE~\cite{dettmers2018convolutional}.
\newcite{scarlini2020more} report the performance of using BERT base-multilingual-cased (mBERT) instead of BERT large with MFS fallback.
\newcite{hadiwinoto2019improved} integrating pretrained BERT model with gated linear unit (GLU) and layer weighted (LW). 

GlossBERT~\cite{huang2019glossbert} fine tunes the pretrained BERT model by jointly encoding contexts and glosses. 
LMMS~\cite{loureiro2019language} learns sense embeddings using BERT to generate contextualised embeddings from semantic networks and sense definitions.
To perform WSD, they use the $1$-NN method and compare sense embeddings against contextualised embeddings generated by BERT.
\newcite{scarlini2020more} augment UKB with SyntagNet’s relations~\cite{scozzafava2020personalized} and obtain UKB$_{+Syn}$.
SensEmBERT is a knowledge-based approach, which produces sense embeddings by means of BabelNet and Wikipedia. 
Although SensEmBERT is effective in modelling nominal meanings, it only consists of nouns due to the limitation of its underlying resources.
SensEmBERT$_{sup}$ is the supervised version of SensEmBERT.
ARES~\cite{scarlini2020more} is a semi-supervised approach for learning sense embeddings by incorporating sense annotated datasets, unlabelled corpora and knowledge bases.

\begin{table}[t]
\centering
\resizebox{0.5\textwidth}{!}{
\begin{tabular}{lcccccc}
\toprule
Models &SE2 &SE3 &SE07 &SE13 &SE15 &ALL \\
\midrule
MFS &65.6 &66.0 &54.5 &63.8 &67.1 &65.6\\
SemCor+OMSTI$_{BERT}$ &74.0 &70.6 &63.1 &72.4 &75.0 &72.2\\
ELMo $k$-NN &71.5 &67.5 &57.1 &65.3 &69.9 &67.9\\
BERT $k$-NN &76.3 &73.2 &66.2 &71.7 &74.1 &73.5\\
EWISE$_{ConvE}$ &73.8 &71.1 &67.3 &69.4 &74.5 &71.8\\
mBERT $k$-NN + MFS &72.7 &70.1 &62.4 &69.0 &72.0 &70.5\\
BERT$_{GLU+LW}$ &75.5 &73.4 &68.5 &71.0 &76.2 &74.0\\
GlossBERT &77.7 &75.2 &\pmb{76.1} &72.5 &80.4 &77.0 \\
LMMS &76.3 &75.6 &68.1 & 75.1 &77.0 &75.4\\
UKB$_{+Syn}$ &71.2 &71.6 &59.6 &72.4 &75.6 &71.5\\
SensEmBERT &70.8 &65.4 &58.0 &74.8 &75.0 &70.1 \\
SenseEmBERT$_{sup}$ &72.2 &69.9 &60.2 &\pmb{78.7} &75.0 &72.8\\
ARES & 78.0 &\underline{77.1} &71.0 &77.3 &\pmb{83.2} &77.9 \\
\midrule
\textit{Proposed Method} \\
CDES$_{linear}$ &\pmb{78.4} & 76.9 &71.0 &77.6 & \underline{83.1} &\underline{78.0}\\
CDES$_{ReLU}$ &\underline{78.1} &\underline{77.1} &71.0 &77.5 &\underline{83.1} &\underline{78.0} \\
CDES$_{GELU}$ &\underline{78.1} &\pmb{77.3} &\underline{71.4} &\underline{77.7} &\pmb{83.2} &\pmb{78.1} \\
\bottomrule
\end{tabular}
}
\caption{F1 scores (\%) for English all-words WSD on the test sets of \protect\newcite{raganato2017word}. Bold and underline indicate the best and the second best results, respectively.}
\label{tbl:wsd}
\end{table}

To study the effect of using a nonlinear mapping $f$ between static and contextualised embedding spaces in \eqref{eq:loss}, we train CDES with linear, ReLU and GELU activations to create respectively CDES$_{linear}$, CDES$_{ReLU}$ and CDES$_{GELU}$ versions.
From Table~\ref{tbl:wsd} we see that among these versions, CDES$_{GELU}$ outperforms the linear and ReLU versions in all datasets, except on SE2 where CDES$_{linear}$ performs best.
This result shows that nonlinear mapping (GELU) to be more appropriate for extracting sense-related information from contextualised embeddings.
Moreover, we see that CDES versions consistently outperform all previously proposed sense embeddings, except on SE07 and SE13 where GlossBERT and SenseBERT$_{sup}$ performs best respectively.
On SE15, the performance of CDES$_{GELU}$ is equal to that of ARES.

Overall, CDES$_{linear}$ obtains the best performance on SE2, while CDES$_{GELU}$ performs best on SE3, SE15 and ALL.
This result provides empirical support to our working hypothesis that contextualised embeddings produced by NLMs encode much more information beyond sense related information, which must be filtered out using $\mat{W}$.
CDES is able to accurately extract the sense-specific information from contextualised embeddings generated by a pretrained NLM to create sense-specific versions of pretrained sense-agnostic static embeddings.

\subsection{Words in Context (WiC)}
\label{sec:wic}

\newcite{Pilehvar:2019} introduced the WiC dataset for evaluating sense embedding methods.
For a particular word $u$, WiC contains pairs of sentences, ($t_{1}$, $t_{2}$) where the same (\emph{positive}) or different (\emph{negative}) senses of $u$ can occur.
An accurate sense embedding method must be able to discriminate the different senses of an ambiguous word.
The problem is formalised as a binary classification task and classification accuracy is reported as the evaluation metric.
A method that assigns the same vector to all of the senses of a word would report a chance-level (i.e. $50\%$) accuracy on WiC.

Similar to Section~\ref{sec:WSD}, we first determine the sense-specific embeddings of $u$, $\vec{s}_{i}(u)$ and $\vec{s}_{j}(u)$ for the senses of $u$ used in respectively $t_{1}$ and $t_{2}$.
We then train a binary logistic regression classifier using the train split of WiC, where we use the cosine similarities between the two vectors in the following six pairs as features, comparing sense and contextualised embeddings in the two sentences.: ($\vec{s}_{i}(u)$, $\vec{s}_{j}(u)$), ($\vec{\zeta}(u,t_{1})$, $\vec{\zeta}(u,t_{2})$), ($\vec{s}_{i}(u)$, $\vec{\zeta}(u,t_{1})$), ($\vec{s}_{j}(u)$, $\vec{\zeta}(u,t_{2})$), ($\vec{s}_{i}(u)$, $\vec{\zeta}(u,t_{2})$) and ($\vec{s}_{j}(u)$, $\vec{\zeta}(u,t_{1})$).
We train this classifier using the official train split in WiC.
In particular, we do not fine-tune the static or contextualised embeddings that are used as inputs by CDES on WiC because our goal is to extract sense-related information already present in the pretrained embeddings.

\begin{table}[t]
\centering
\small 
\begin{tabular}{lc}
\toprule
Models &Accuracy \% \\
\midrule
\textit{Static Embeddings} \\

GloVe~\cite{pennington2014glove} & 50.9\\
\midrule
\textit{Contextualised Embeddings} \\
ElMo~\cite{peters2018deep} &57.7\\
ELMo-weighted~\cite{ansell2019elmo} &61.2\\
BERT-large~\cite{devlin2019bert} &65.5\\
RoBERTa~\cite{liu2019roberta} &69.9\\
KnowBERT-W+W~\cite{peters2019knowledge} &70.9\\
SenseBERT-large~\cite{levine-etal-2020-sensebert} &\underline{72.1} \\
BERT$_{ARES}$~\cite{scarlini2020more} &\pmb{72.2} \\
\midrule
\textit{Static Sense Embeddings} \\
MUSE~\cite{lee2017muse} &48.4 \\
LMMS~\cite{Loureiro2019LIAADAS} &67.7\\
LessLex~\cite{colla-etal-2020-lesslex} &59.2 \\
CDES$_{linear}$ &69.0 \\
CDES$_{ReLu}$ &68.6 \\
CDES$_{GELU}$ &68.8\\

\bottomrule
\end{tabular}
\caption{Performance on WiC. Bold and underline respectively indicate the best and the second best results.}
\label{tbl:wic-dev}
\end{table} 

\begin{figure*}[t]
\centering
\begin{minipage}[t]{0.33\linewidth}
\centering
\includegraphics[width=2.0in, height=2.0in]{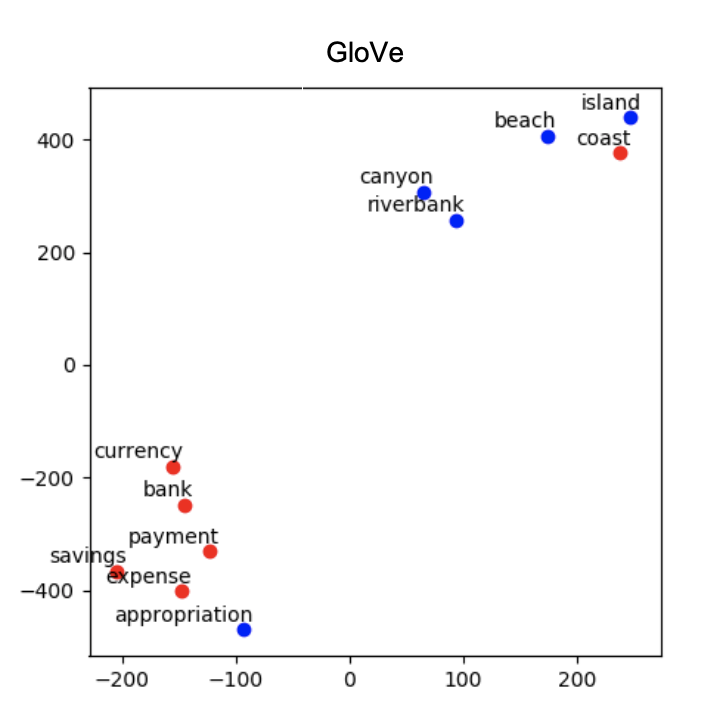}
\label{fig:side:a}
\end{minipage}%
\begin{minipage}[t]{0.33\linewidth}
\centering
\includegraphics[width=2.0in, height=2.0in]{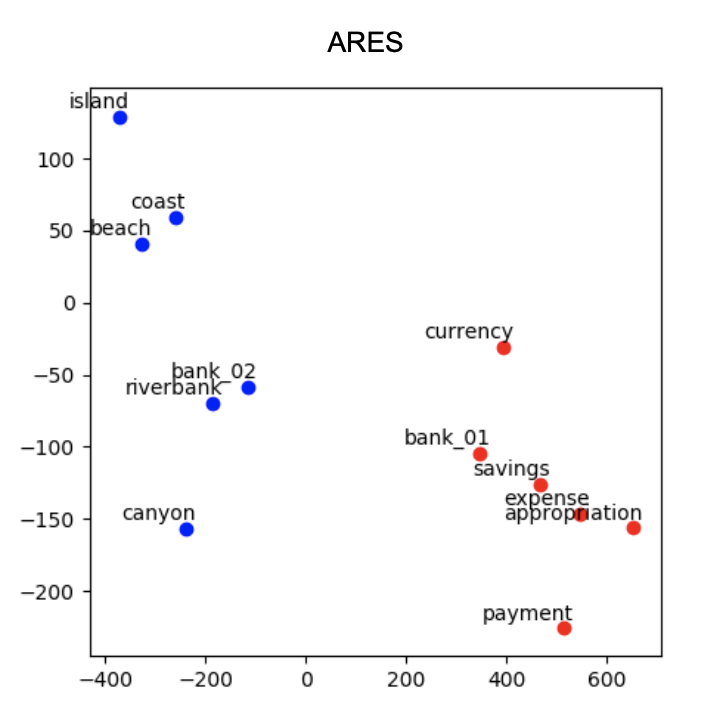}
\end{minipage}
\begin{minipage}[t]{0.33\linewidth}
\centering
\includegraphics[width=2.0in, height=2.0in]{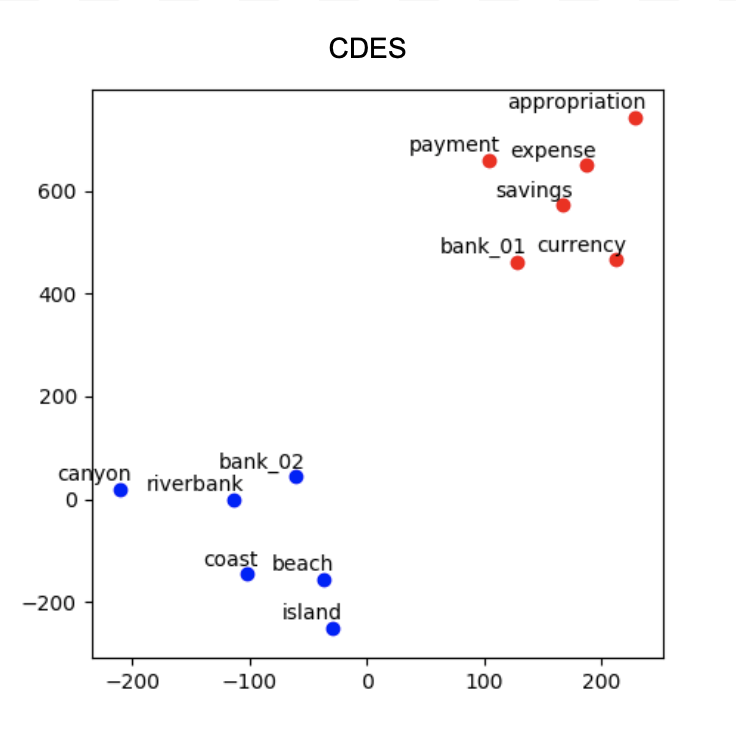}
\end{minipage}
\caption{t-SNE visualisations of the nearest neighbours of \emph{bank} corresponding to the two senses  \emph{financial institution} (in red) and \emph{sloping land} (in blue) are shown for GloVe, ARES and CDES embeddings. Sense labels of synonyms are omitted to avoid cluttering.}
\label{fig:tsne_bank}
\end{figure*} 

In Table~\ref{tbl:wic-dev}, we report the classification accuracies on WiC  for different types of embeddings such as 
static word embeddings (GloVe), contextualised embeddings generated by NLMs (ELMo, ElMo-weighted, BERT-large, RoBERTa and KnowBERT), and sense-specific embeddings (MUSE, LMMS, LessLex, SenseBERT-large and BERT$_{ARES}$).
Due to space limitations we omit the details of these embeddings.

From Table~\ref{tbl:wic-dev} we see that SenseBERT-large and BERT$_{ARES}$ obtain better performance than other embeddings.
All the CDES variants outperform previous static sense embeddings learning methods.
However,  MUSE\footnote{\url{https://github.com/MiuLab/MUSE}} do not assign sense labels to sense embeddings as done by LMMS, LessLex and CDES.
Among CDES variants, CDES$_{linear}$ performs best and is closely followed by GELU and ReLU variants.
Although, CDES variants do not surpass the current SoTA methods such as SenseBERT-large and BERT$_{ARES}$ on WiC, unlike CDES these methods fine-tune on WiC train data and/or use more complex classifiers with multiple projection layers compared to the single logistic regression over six features used by CDES.\footnote{BERT$_{ARES}$ and SenseBERT use respectively 2048 and 1024 features for sense prediction in WiC.}
More importantly, results from both WSD and WiC experiments support our claim that contextualised embeddings encode word sense related information that can be extracted and injected into sense-insensitive static word embeddings via (non)linear projections to create sense-sensitive versions of the sense-insensitive static embeddings. 






\subsection{Visualisation of Sense Embeddings}
\label{sec:nonlinear}

To visualise the embeddings corresponding to the different senses of an ambiguous word, we consider \emph{bank}, which has the two distinct senses: \emph{financial institution} and \emph{sloping land}.
We randomly select $5$ synonyms for each sense of \emph{bank} from the WordNet and project their sense/word embeddings using  t-SNE in Figure~\ref{fig:tsne_bank}.
Compared to GloVe, we see that words with related meanings are projected onto coherent clusters by ARES and CDES. 
This indicates that sense embeddings are able to distinguish polysemy correctly compared to static word embeddings. 
Overall, we see that CDES produces better separated clusters than both GloVe and ARES.

\subsection{Nearest Neighbours of Sense Embeddings}
\label{sec:NN}

\begin{table*}[t!]
\centering
\small
\begin{tabular}{p{2.7cm}p{2.7cm}p{2.7cm}p{2.7cm}p{2.7cm}}
\toprule
\multicolumn{5}{p{16cm}}{\textbf{Sentence 1:} The \colorbox{cyan}{banks} which held the mortgage on the old church declared that the interest was considerably in arrears, and the real estate people said flatly that the land across the river was being held for an eventual development for white working people who were coming in, and that none would be sold to colored folk.} \\
\midrule
GloVe & BERT & LMMS & SenseBERT & CDES\\
\midrule 
mortgage & mortgage & mortgage & mortgage & mortgage \\ 
interest & interest & church & real & real estate \\
estate & held & sell & old & sell \\
river & church & interest & land & interest \\
real & river & real estate & interest & church \\
\midrule \midrule
\multicolumn{5}{p{16cm}}{\textbf{Sentence 2:} Through the splash of the rising waters, they could hear the roar of the river as it raged through its canyon, gnashing big chunks out of the \colorbox{red}{banks}.} \\
\midrule
GloVe & BERT & LMMS & SenseBERT & CDES\\
\midrule 
mortgage & river & river & splash & river \\
interest & waters & canyon & land & water \\
estate & chunks & land & out & rise \\
river & splash & folk & through & canyon \\
real & canyon & church & chunks & folk \\
\bottomrule
\end{tabular}
\caption{Nearest neighbours computed using the word/sense embeddings of \emph{bank} in two sentences.}
\label{tbl:NN}
\end{table*}

An accurate sense embedding method must be able to represent an ambiguous word with different embeddings considering the senses expressed by that word in different contexts.
To understand how sense embedding of a word vary in different contexts, we compute the nearest neighbours of an ambiguous word using its sense embedding.
Table~\ref{tbl:NN} shows two sentences from SemCor containing \emph{bank}, where in Sentence 1, \emph{bank} takes the \emph{financial institution} sense, and in Sentence 2 the \emph{sloping land (especially the slope beside a body of water)} sense.
We compute the sense embedding of \emph{bank}, given each sentence as the context, using different methods and compute the top 5 nearest neighbours, shown in the descending order of their cosine similarity scores with the sense embedding of \emph{bank} in each sentence.

GloVe, which is sense and context insensitive uses the same vector to represent \emph{bank} in both sentences, resulting in the same set of nearest neighbours, which is a mixture of finance and riverbank related words.
On the other hand, BERT, which is context-sensitive but not sense-specific, returns different sets of nearest neighbours in the two cases.
In particular, we see that finance-related nearest neighbours such as \emph{mortgage} and \emph{interest} are selected for the first sentence, whereas riverbank-related nearest neighbours such as \emph{water} and \emph{canyon} for the second. 
However, BERT does not provide sense embeddings and some neighbours such as \emph{river} appear in both sets, because it appears in the first sentence, although not related to \emph{bank} there.

SenseBERT~\cite{levine-etal-2020-sensebert} disambiguates word senses at a coarse-grained WordNet's supersense level.
We see that SenseBERT correctly detects words such as \emph{mortgage} and \emph{interest} as neighbours of \emph{bank} in the first sentence, and \emph{splash} and \emph{land} in the second.
We see that \emph{land} appears as a nearest neighbour in both sentences, although it is more related to the \emph{sloping land} sense than the \emph{financial institution} sense of \emph{bank}.

LMMS selects \emph{church} as a nearest neighbour for both sentences, despite being irrelevant to the second.
On the other hand, CDES correctly detects \emph{church} for the first sentence and not for the second.
Overall, CDES correctly lists financial institution sense related words such as \emph{mortgage}, \emph{real estate} and \emph{interest} for the first sentence, and sloping land sense related words such as \emph{river}, \emph{water} and \emph{canyon} in the second sentence.

%


\section{Related Work}

\newcite{reisinger2010multi} proposed multi-prototype embeddings to represent word senses, which was extended by \newcite{huang2012improving} combining both local and global contexts.  
Both methods use clustering to group contexts of a word related to the same sense.
Although the number of senses depends on the word, these methods assign a fixed number of senses to all words.
To overcome this limitation, \newcite{neelakantan2014efficient} proposed a non-parametric model, which estimates the number of senses dynamically per word.

Even though clustering-based methods are able to assign multi-prototype embeddings for a word, they still suffer from the fact that the trained embeddings are not associated with any sense inventories~\cite{Camacho_Collados_2018}.
In contrast, knowledge-based approaches learn sense embeddings by extracting sense-specific information from external sense inventories, such as WordNet and BabelNet.
\newcite{chen2014unified} extended word2vec~\cite{mikolov2013distributed} to learn sense-specific embeddings associated with the WordNet~\cite{miller1998wordnet} \emph{synsets}.
\newcite{rothe2015autoextend} used the semantic relations in WordNet to embed words and their senses into a common vector space.  
\newcite{iacobacci2015sensembed} use the sense definitions from BabelNet and perform word sense disambiguation (WSD) to obtain sense-specific contexts. 

Recently, contextualised embeddings generated by NLMs have been used to create sense embeddings.
\newcite{loureiro2019language} construct sense embeddings by taking the average over the contextualised embeddings of the sense annotated tokens from SemCor.
\newcite{scarlini2020sensembert} propose a knowledge-based approach for constructing BERT-based embeddings of senses by means of the lexical-semantic information in BabelNet and Wikipedia.
CDES proposed in this paper extends this line of work, where we extract sense-related information from an NLM and inject into a static word embedding to create a sense-specific version of the latter.
Moreover, we follow prior work~\cite{scarlini2020more,loureiro2019language} and incorporate contexts from external resources such as Wikipedia, WordNet and SyntagNet representing different senses of a word to enhance sense embeddings learnt using sense-labelled corpora.

\section{Conclusion}
We proposed CDES, a method which is able to generate sense embeddings by extracting the sense-related information from contextualised embeddings. 
CDES integrates the gloss information from a semantic network as well as the information from an external corpus to tackle the sense-coverage problem.
Evaluations on multiple benchmark datasets related to WSD and WiC tasks show that CDES learns accurate sense emebddings,
and report comparable results to the current SoTA.
All experiments reported in the paper are limited to English language and we plan to extend the proposed method to learn multilingual sense embeddings in our future work.




\bibliography{sense.bib}
\bibliographystyle{acl}
\end{document}


\maketitle

\section{Experimental Settings and Hyperparameters}

We use Adam~\cite{Kingma:ICLR:2015} as the optimiser and set the minibatch size to $64$ with an initial learning rate of 1E-4. 
GloVe is used as the pre-trained static word embedding and contain $300$ dimensional embeddings, whereas BERT-large model is used as the pre-trained contextualised word embeddings and $1024$ dimensional.
In order to map the contextualised embeddings and static embeddings in to a common vector space, we set $\mat{W} \in \R^{300 \times 1024}$.
To reduce the memory footprint, number of trainable parameters and thereby overfitting, we constrain the sense-specific matrices $\mat{A}_i \in \R^{300 \times 300}$ to be diagonal matrices.
$\mat{W}$ and $\mat{A}_{i}$ matrices are randomly initialised following Xavier initialisation~\cite{Glorot:AISTAT:2010}.
Following the work of~\newcite{wiedemann2019does}, we set $k=3$ for the WSD task, which means that we consider three potential senses when conducting WSD task.
For WiC task we set $k=1$ for all comparisons.
All the above hyperparameter values were tuned using a randomly selected subset of training data as validation data.

In terms of visualisation, we use \texttt{sklearn.manifold.TSNE} to perform t-SNE visualisation. 
We set the n\_components=2, init=`pca', perplexity=3, n\_iter=1500 and metric=`cosine'.

\bibliography{sense.bib}
\bibliographystyle{acl_natbib}